\pdfoutput=1

\documentclass[11pt]{article}
\usepackage[final]{acl}

\usepackage{times}
\usepackage{graphicx}
\usepackage{makecell}
\usepackage{multirow}
\usepackage{multicol}
\usepackage{colortbl}
\usepackage[T1]{fontenc}
\usepackage[utf8]{inputenc}
\usepackage{microtype}
\usepackage{inconsolata}
\usepackage{xspace}
\usepackage{amsmath}
\usepackage{diagbox}
\usepackage{booktabs}
\usepackage{wrapfig}
\usepackage{listings}
\usepackage{svg}
\usepackage{rotating}
\usepackage{pdfpages}
\usepackage{enumitem}
\usepackage[capitalize,noabbrev]{cleveref}
\usepackage{comment}
\usepackage{dblfloatfix}
\usepackage{float}
\usepackage{caption}
\usepackage[capitalize,noabbrev]{cleveref}
\usepackage[most]{tcolorbox}
\usepackage{xcolor,colortbl}
\usepackage{float}

\definecolor{softgreen}{RGB}{34,139,34}

\newcommand{\corpus}{\texorpdfstring{\textsc{LexTime}}\xspace}

\title{\corpus: A Benchmark for Temporal Ordering of Legal Events}

\author{
Claire Barale$^1$ \qquad Leslie Barrett$^2$ \qquad Vikram Sunil Bajaj$^2$ \qquad Michael Rovatsos$^1$\\[0.5em]
$^1$School of Informatics, University of Edinburgh \qquad
$^2$Bloomberg\\[0.3em]
\small{
\texttt{\{claire.barale, Michael.Rovatsos\}@ed.ac.uk} \hspace{3em}
\texttt{\{lbarrett4, vbajaj11\}@bloomberg.net}
}
}

\begin{document}
\maketitle
\begin{abstract}
Understanding temporal relationships and accurately reconstructing the event timeline is important for case law analysis, compliance monitoring, and legal summarization. However, existing benchmarks lack specialized language evaluation, leaving a gap in understanding how LLMs handle event ordering in legal contexts. We introduce \corpus, a dataset designed to evaluate LLMs' event ordering capabilities in legal language, consisting of 512 instances from U.S. Federal Complaints with annotated event pairs and their temporal relations\footnote{\url{https://zenodo.org/records/17157439}}. Our findings show that (1) LLMs are more accurate on legal event ordering than on narrative texts (up to +10.5\%); (2) longer input contexts and implicit events boost accuracy, reaching 80.8\% for implicit-explicit event pairs; (3) legal linguistic complexities and nested clauses remain a challenge. While performance is promising, specific features of legal texts remain a bottleneck for legal temporal event reasoning, and we propose concrete modeling directions to better address them. 
\end{abstract}

\section{Introduction} \label{sec:intro}
Understanding the temporal relationships between events is fundamental for natural language understanding. In the legal domain, temporal reasoning is important for tasks such as case law analysis, contract interpretation, and compliance monitoring. Accurately determining the timeline of events helps assess liability, obligations, and procedural validity. However, the unique linguistic characteristics of legal texts raise questions about the effectiveness of current temporal reasoning approaches.

Language models (LMs) face notable challenges with temporal reasoning due to the complex logic and ambiguities involved, and the necessity to process long contextual dependencies \cite{jain-etal-2023-language-models}. While several benchmarks exist for temporal reasoning with large language models (LLMs), none focus on specialized languages. Most current research on event extraction and ordering focuses on narratives, such as the \textsc{Event StoryLine Corpus} \cite{caselli2018events}, the  \textsc{Fine Grained Temporal Relations Dataset} \cite{vashishtha2019fine}, and the \textsc{Tracie} dataset \cite{zhou-etal-2021-temporal}. As a result, LMs' proficiency in expert types of language, such as legal language, is largely unknown. How do LMs address the unique features of legal language for event ordering? 

\begin{figure}[t]
\includegraphics[width=\columnwidth]{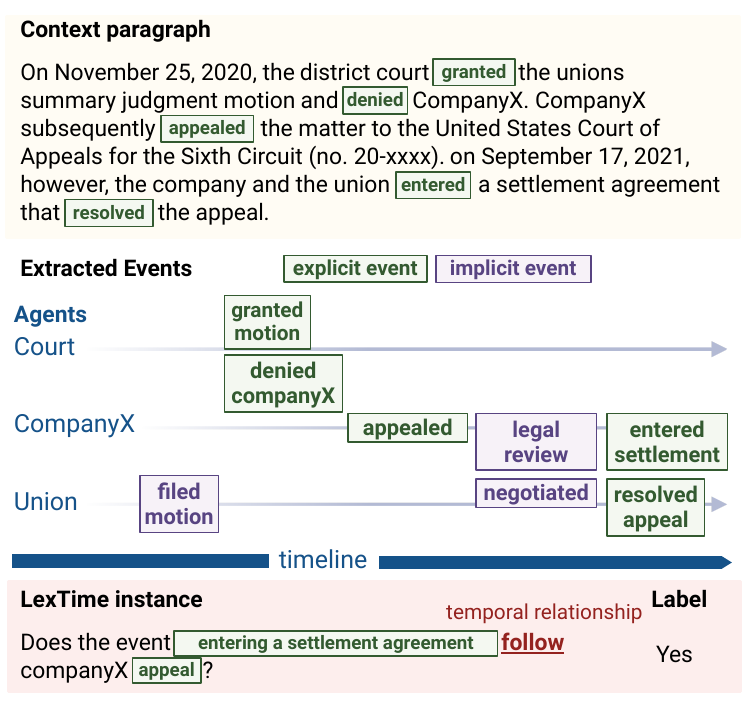} 
\centering
   \caption{Legal event ordering task in \corpus: explicit and implicit events (\textsc{middle}) are extracted from a context paragraph (\textsc{top}) and combined into a yes/no question (\textsc{bottom}).}
   \label{figure: task_overview}
\end{figure}

In this work, we conduct the first study targeting temporal event ordering within legal language. Besides its implications for legal work, event ordering is a foundation for downstream legal NLP tasks, such as summarization, drafting, and question-answering. We start by assembling \corpus, a novel dataset composed of 512 instances constructed from labor-related U.S. Federal Complaints (\S \ref{sec:dataset}). We extract events relevant to both the case and its procedural timeline, providing a structured resource for studying temporal reasoning in legal texts. An example instance from \corpus\space is shown in Fig. \ref{figure: task_overview}.

We show that legal language used in \corpus\space  exhibits specific lexical, syntactic, and discourse characteristics, making it a distinct genre (\S \ref{sec:linguistic_analysis}). Since event temporal ordering has been extensively studied in narrative contexts, we use \textsc{Tracie} as a baseline for comparison due to its similarity in task design and focus on event ordering. 

Next, we conduct a systematic evaluation of LMs' ability to perform event ordering, defined as identifying temporal relations between event pairs (\S \ref{sec:methodo}, \ref{sec:results}). Our analysis focuses on key aspects relevant to the legal domain, such as the impact of context length, the distinction between explicit and implicit events, and a comparative study of narrative versus legal language. Following \citet{zhou2021temporalreasoningimplicitevents}, we define an implicit event as one whose trigger is not explicitly stated in the text but can be inferred from the surrounding context, such as \textit{negotiated} in Fig. \ref{figure: task_overview}. A key indicator of the implicit event is that it can be fully entailed from and properly implied by the following explicitly stated events.

Our experiments assess Chain-of-Thought (CoT) prompting \cite{10.5555/3600270.3602070}, a method previously successful in reasoning over complex and long inputs. However, we find that CoT prompting is ineffective for event ordering, leading to a 3.4\% accuracy drop for \corpus\space compared to standard few-shot prompting, which yields the best accuracy.

Overall, we have three core findings: (1) LLMs show improved performance on legal event ordering in \corpus\space compared to narrative, achieving enhancements of up to +10.5\% (GPT-4o, few-shot). This suggests that legal texts offer clearer temporal structures for models to use. (2) LLMs perform better with longer input contexts (+2.3\% for GPT-4 Turbo, few-shot) and with event pairs that include implicit event types. The highest accuracy of 80.8\% is achieved when queries contain a pair with one explicit and one implicit event (GPT-4 Turbo, few-shot), surpassing pairs of two explicit events by 2.6\% and the overall dataset average by 3.2\% (\S \ref{sec:results}). (3) However, linguistic complexities commonly found in legal language contribute to model errors (\S\ref{sec:errors_analysis}). Features typical of the legal genre are overrepresented in misclassified instances, such as paraphrases (+89\% in error samples) and events occurring in subordinate clauses (+84\%).

In sum, we make the following contributions:
\begin{enumerate}[leftmargin=*] 
\item We introduce \corpus, the first dataset curated for legal event ordering, enabling the systematic study of temporal reasoning in legal texts.
\item We use \corpus\space to analyze the abilities of large LLMs on legal event ordering, a fundamental task in legal NLP. Our findings reveal that while LLMs achieve higher accuracy on legal texts than narrative texts, their performance is still constrained by linguistic complexities found in legal language.
\item We identify effective strategies for improving model accuracy, showing that longer context windows enhance accuracy for large models. In comparison, shorter context windows benefit smaller models, providing practical insights for optimizing model configurations in legal NLP tasks.
\end{enumerate}

\section{Datasets} \label{sec:dataset}
\corpus{} is a dataset focused on temporal reasoning in legal language, derived from U.S. federal complaints between 2020 and 2024. We randomly sample complaints categorized under the Nature of Suit (NOS) codes beginning with 7, which correspond to labor-related cases\footnote{Complaints are publicly accessible at \url{https://pacer.uscourts.gov/}.}. For comparison, we use \textsc{Tracie} \cite{zhou2021temporalreasoningimplicitevents}, which consists of short stories and has been curated for the same task, offering a natural choice.

\subsection{Task Overview and \corpus\xspace Dataset} \label{subsec:task_dataset}
\corpus\xspace is a dataset built to measure the ability of LMs to perform temporal event ordering on a specific type of language: legal language. Each instance contains:
(1) A context paragraph taken from a complaint document.
(2) A pair of events, which can be either one implicit event and one explicit event or two explicit events. An event is an occurrence or action triggered by a verb or noun that takes place at a specific time and can be ordered relative to other events based on temporal relationships.
(3) A query consisting of a pair of events and a temporal relationship between them, creating a triplet following this template: \textit{\{event A, TR, event B\}}.
(4) A binary label, \textit{yes} or \textit{no}.
For instance, in the positive instance shown in Fig. \ref{figure: task_overview}, \textit{entered a settlement agreement} and \textit{CompanyX appealed} are two explicit events taken from the \textit{context paragraph}, \textit{follow} is the \textit{TR} that links them in the \textit{query}, and the answer (\textit{label}) to the \textit{query} according to the \textit{context} is \textit{yes}.

\paragraph{Pre-labeling.} 
We used \textit{Mistral-Large-Instruct-2407} to help construct instances and then manually reviewed all generated instances (prompts in Appendix \ref{app: prompts_dataset_generation}). We employed a prompt-chaining approach to minimize errors that could occur during the multiple steps involved \cite{wu2022promptchainerchaininglargelanguage}.

\begin{figure}[t]
\includegraphics[width=\columnwidth]{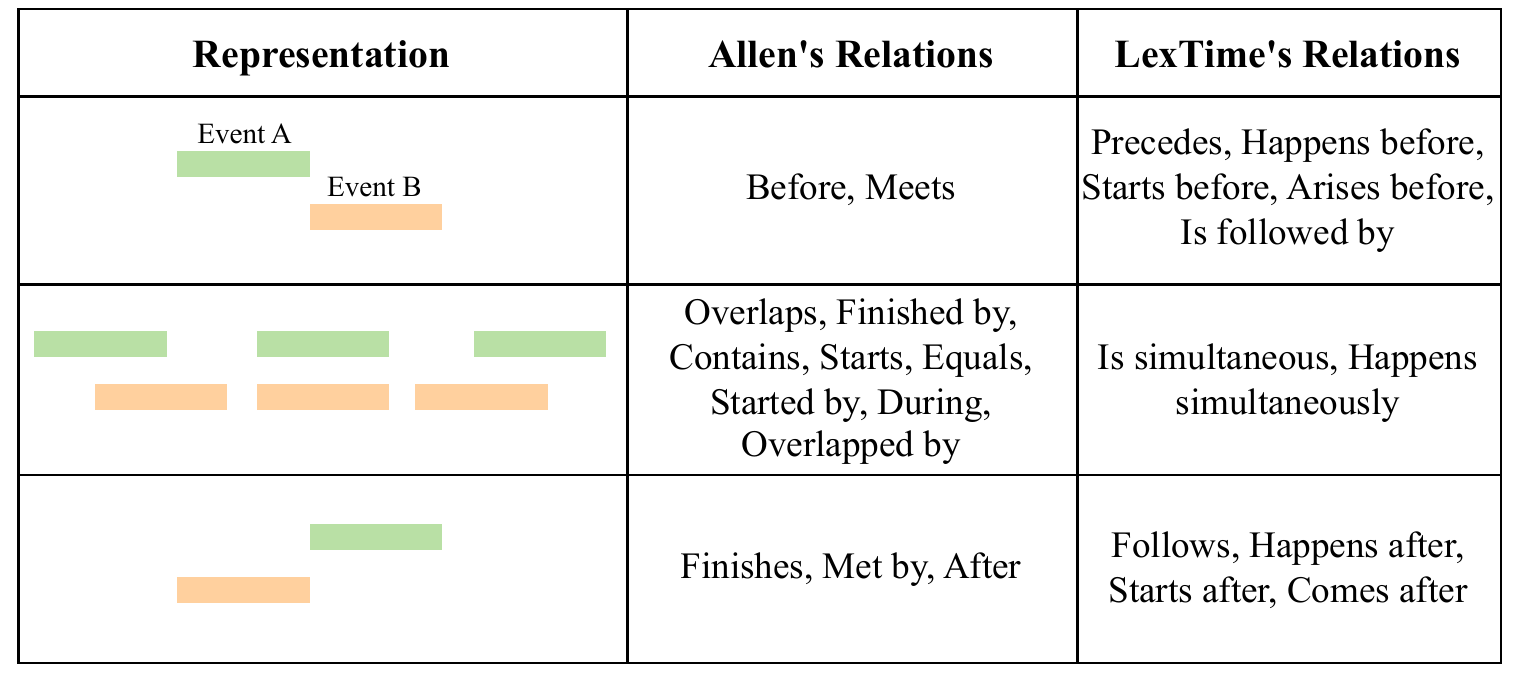} 
\centering
   \caption{\corpus's temporal relations and their correspondence to Allen's interval relations \cite{allen1983maintaining}.}
   \label{figure: allens_relations}
\end{figure}

We kept paragraphs with at least two \textit{-ed} tokens (event triggers). We merged every three paragraphs for more context, resulting in an average \textit{context paragraph} length of 172 tokens. Next, we extracted explicitly mentioned events using an initial prompt. We manually reviewed these to select the most significant case or procedural events. Intermediate cases with ambiguous temporal relations were resolved during this annotation stage through manual review. We then identified five implicit events per paragraph and generated sentences linking events with temporal relationships, indicating whether one event occurred \textit{before, at the same time}, or \textit{after} another. All possible TR variations are provided in the last column of Fig.~\ref{figure: allens_relations}. Each TR was reversed to create a contradictory query (e.g., \textit{happens after} to \textit{happens before}).

Fig. \ref{figure: allens_relations} compares the TR used in \corpus\space with Allen’s relations \cite{allen1983maintaining}, which are foundational to temporal reasoning. We simplify the TR to three categories for practical and theoretical reasons: (1) Strict event order: in legal reasoning, important TRs primarily depend on precedence and causality (e.g., \textit{The lawsuit was filed before the hearing}), and finer-grained relations like \textit{Meets} are less relevant. (2) Ambiguous boundaries: Allen's relations need precise interval boundaries, which are often vague. Simplifying these TRs improves interpretability and consistency and reduces annotation errors. (3) Consistency with  \textsc{Tracie}: we employ three key relationships to ensure alignment and facilitate comparison with \citet{zhou2021temporalreasoningimplicitevents}.

\paragraph{Label collection.} 
We manually reviewed the 512 pre-labeled instances and corrected their labels if needed. In total, we found that 23.24\% of the samples were wrongly labeled. Among those errors, 54.55\% were an error on the temporal relationship, 37.87\% on the event extraction, and 7.58\% on both. We collect a balanced dataset of 257 instances labeled \textit{yes} (TR is correct) and 255 instances labeled \textit{no} (TR is incorrect). 

\subsection{\textsc{\textsc{Tracie}} Dataset}\label{subsec:tracie}
We randomly selected 512 instances from the \textsc{Tracie} dataset \cite{zhou2021temporalreasoningimplicitevents} to match the number of samples in \corpus. \textsc{Tracie} contains short story paragraphs and implicit events. We modified the original \textsc{Tracie} dataset to match the format of \corpus\space by converting instances to a query instead of the original entailment task format. We rerun all experiments as previous benchmarks using \textsc{Tracie} have employed older models \cite{chu-etal-2024-timebench}.


\begin{figure*}[!ht]
\includegraphics[width=0.9\textwidth]{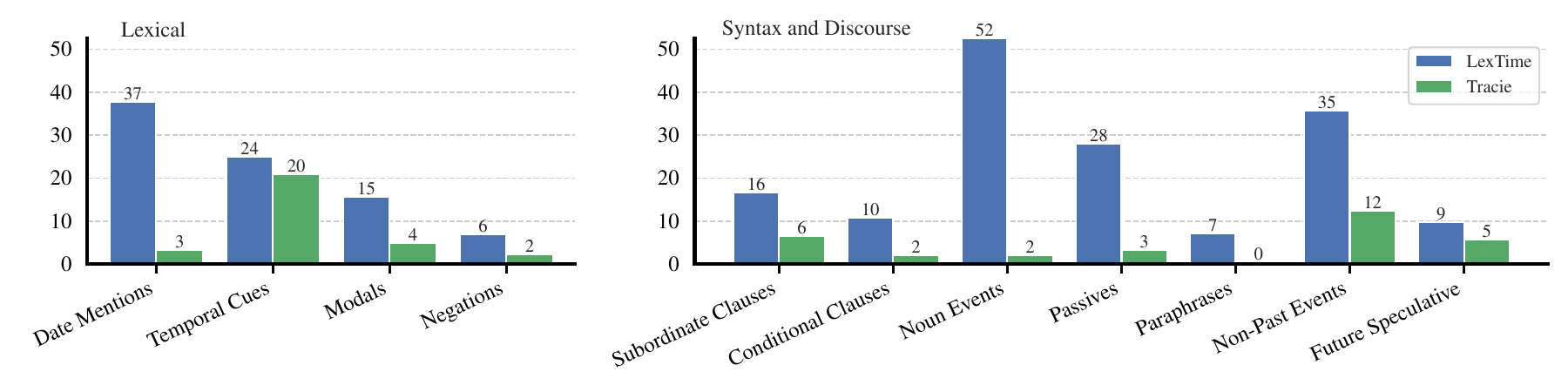} 
\centering
   \caption{Comparison of linguistic features in \corpus\space and \textsc{Tracie}. Values (y-axis) represent the percentage of sentences that contain each feature, indicating its frequency in the text.}
   \label{figure: language_analysis}
\end{figure*}

\begin{figure}[t]
\includegraphics[width=0.9\columnwidth]{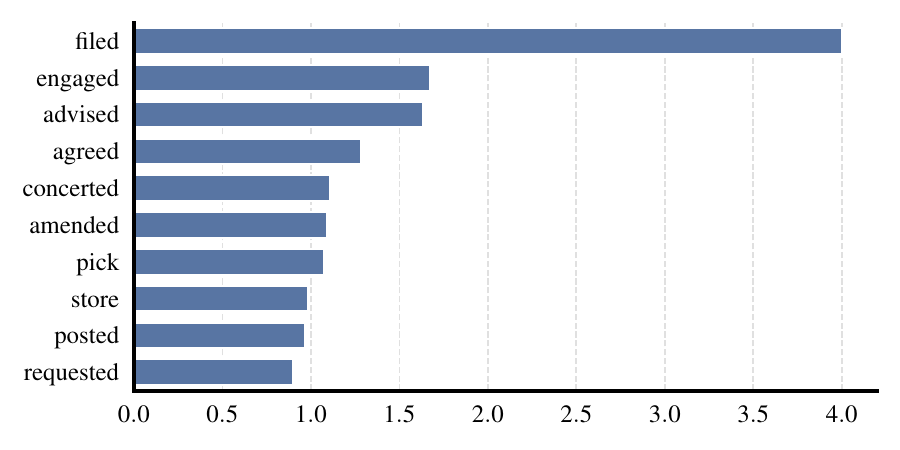} 
\centering
   \caption{Top-10 event triggers in \corpus{} by frequency (\%), showing specialized legal vocabulary.}
   \label{figure: topevent_triggers}
\end{figure}

\section{Specific Features of Legal Events} \label{sec:linguistic_analysis}
This section examines the linguistic and structural characteristics that differentiate legal language from narrative, using \corpus\space and \textsc{Tracie} as representative datasets. We show that legal language constitutes a distinct genre characterized by specialized event structures and temporal representations.

Prior work has shown that legal language follows distinct linguistic conventions \cite{yunus2016colligations}, including lengthy nominalized expressions, frequent passive forms, and specialized vocabulary, making its syntactic structure more complex than that of general prose or news texts \cite{kuzman2023automatic}. In an empirical study of the text in the U.S. Code, \citet{martinezmollica2024} found higher rates of complex syntactic structures relative to six baseline genres of English. Similarly, \citet{martinezmollica2022} noted that U.S. case law contains center-embedded clauses at double the rate of other text genres. Guided by the features identified by \citet{yunus2016colligations}, we divide our analysis into: (1) a lexical analysis and (2) a syntax and discourse analysis (Fig. \ref{figure: language_analysis}).

We employ a combination of methods, each selected for its suitability to a specific feature of interest: (1) named-entity recognition to extract dates; (2) a set of rules and keyword search to extract temporal markers (Appendix \ref{app: temp_keywords}); (3) part-of-speech and dependency parsing to extract modal verbs, tense of the verbs, negations; (4) we further the dependency parsing with manual annotation for nominalized events, passive voice, citations, subordinate and conditional clauses.

\subsection{Vocabulary}
\paragraph{Domain-specific event triggers.}

An event trigger is a verb or noun that signals an event, such as \textit{filed} in \textit{filed a lawsuit}. In \corpus, event triggers are mostly legal terms (Fig. \ref{figure: topevent_triggers}), and they vary significantly from those in \textsc{Tracie} (Appendix \ref{app: tracie_triggers}). None of the 10 most frequent event triggers overlap between the datasets, and we compute a low lexical overlap of 0.09 using all unique event triggers found in each dataset (Jaccard similarity). \textsc{Tracie} shows a higher diversity of event triggers with a ratio of 0.148 and 701 unique triggers, compared to 0.035 and 403 unique triggers for \corpus\space (unique triggers divided by the total number of triggers). 

Many terms in \corpus\space have specific meanings that differ from their general usage (semantic shift). For example, \textit{charged} in legal terms means a formal accusation, while in everyday language, it can refer to \textit{charging a phone}. In contrast, \textsc{Tracie} includes event triggers that often describe everyday actions (e.g., \textit{started, asked}).



\paragraph{Precise and explicit temporal cues.}
\corpus\space contains 10.28x more date mentions than \textsc{Tracie}, reflecting reliance on precise and explicit temporal expressions (e.g., \textit{on June 4, 2021}) and statutory deadlines (e.g., \textit{within 30 days of filing}). In contrast, narrative texts use vague temporal expressions, as evidenced by fewer dates but a higher occurrence of frequency adverbs (+2.43\%, Appendix \ref{app: chart_temp_adverbs}).

\paragraph{Higher density of temporal markers.}
\corpus{} contains a higher density of temporal markers than \textsc{Tracie} (+3.99\%), highlighting a structural and functional difference in the representation of events (Fig. \ref{figure: language_analysis}).

\paragraph{Frequent use of modalities.}
\corpus\space has 11\% more modal verbs than \textsc{Tracie}, except for \textit{could}, which shows a decrease of -1.96\% (Appendix \ref{app: chart_modals}). The use of modalities in legal writing indicates binding obligations (\textit{must, shall}), permissions (\textit{may, can}), and advisory language (\textit{should}). In contrast, narratives employ modality to express uncertainty (\textit{could}). 


\paragraph{Frequent use of negation.}
Negation appears 4.73\% more frequently in \corpus\space than in \textsc{Tracie}. This reflects its structural and functional role in legal reasoning (restrictions, prohibitions, and exceptions). \corpus\space includes double negatives, characteristic of legal discourse. For example: 
\textit{A decision shall not be invalid solely because of a procedural error}. 

\subsection{Syntax, Discourse, and Tenses}

\paragraph{Events in subordinate or conditional clauses.}
Subordinate and conditional clause events are more frequent in \corpus\space than in \textsc{Tracie} (+10.5\% and +9.56\%). This reflects complex hierarchical dependencies, where an event's occurrence can depend on another event or condition.

\begin{figure}[H]
\includegraphics[width=0.95\columnwidth]{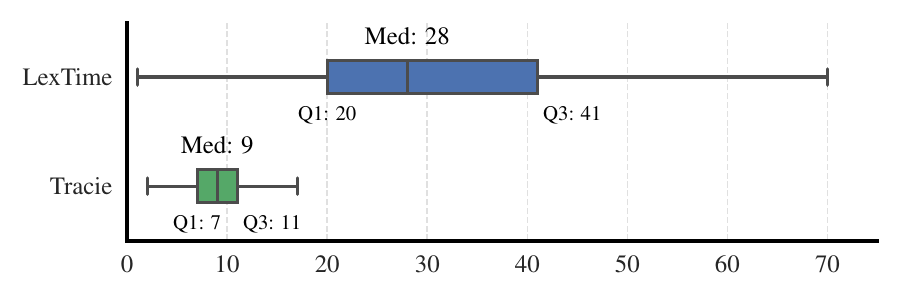} 
\centering
   \caption{Boxplot showing sentence length distributions (in tokens). Median (\textit{Med}) and quartiles (\textit{Q1, Q3}) are noted, outliers are excluded for clarity.}
   \label{figure: boxplot}
\end{figure}

\paragraph{Sentence length and event density.} 
\corpus\space has longer sentences than \textsc{Tracie} (Fig. \ref{figure: boxplot}), averaging 34.72 vs. 8.86 tokens, resulting in higher event density (4.01 vs. 1.60 events per sentence). 

\paragraph{Nominalization, passive voice, and paraphrasing.}
Legal writing often uses nouns instead of verbs to express events, for instance: \textit{The revocation of the contract was issued yesterday}. Nominalized events appear in 52.50\% of sentences (+50.4\% compared to \textsc{Tracie}). Legal texts prioritize nominalization to emphasize outcomes and obligations. They also use the passive voice (+24.76\%) and paraphrase reported events (+6.56\%) more frequently, enhancing their impersonal and authoritative tone. Passives create implicit event structures by omitting the agent. For example, in \textit{The revocation of the contract}, the actor is unstated, making it difficult to identify who acted and when. 

\begin{table*}[h!]
\centering
\resizebox{0.75\textwidth}{!}{
\begin{tabular}{lccc|cccccc|cccccc}
\toprule

\textbf{Setup} & \multicolumn{3}{c|}{\textbf{ All }} & \multicolumn{3}{c}{\textbf{ Long Context }} &  \multicolumn{3}{c|}{\textbf{ Short Context }}  & \multicolumn{3}{c}{\textbf{ Explicit-Explicit }}&  \multicolumn{3}{c}{\textbf{ Explicit-Implicit }} \\

& ZS & 1S & FS & ZS & 1S & FS & ZS & 1S & FS & ZS & 1S & FS & ZS & 1S & FS \\  
\midrule

GPT-4o & 69.9 & \cellcolor{blue!30}74.9 & \cellcolor{blue!40}77.4 & \cellcolor{blue!10}70.3 & \cellcolor{blue!30}74.6 & \cellcolor{blue!40}77.4 & \cellcolor{blue!10}71.6 & 71.2 & \cellcolor{blue!40}77.0 & 68.8 & \cellcolor{blue!20}74.8 &\cellcolor{blue!40} 76.8 & \cellcolor{blue!10}71.5 & \cellcolor{blue!30}76.9 & \cellcolor{blue!40}79.0\\
\textit{   +CoT}  & - & 68.6 & \cellcolor{blue!20}72.6 & - & 69.0 & \cellcolor{blue!20}72.6 & - & 68.7 & \cellcolor{blue!20}73.2 & - & 69.3 & \cellcolor{blue!10}72.7 & - & 69.2 & \cellcolor{blue!20}74.0 \\
\midrule

GPT-4 Turbo & 70.2 & \cellcolor{blue!30}75.1 &\cellcolor{blue!40} 77.2 & 70.1 & \cellcolor{blue!30}76.6 & \cellcolor{blue!40}79.9 & 70.1 & \cellcolor{blue!30}73.5 & \cellcolor{blue!40}76.4 & 71.0 & \cellcolor{blue!40}76.6 & \cellcolor{blue!40}78.2 & \cellcolor{blue!10}71.1 & \cellcolor{blue!30}74.3 &\cellcolor{blue!40} 80.8 \\  
\textit{   +CoT}  & - & \cellcolor{blue!40}77.6 & \cellcolor{blue!30}74.3 & - & \cellcolor{blue!40}77.7 & \cellcolor{blue!20}71.3 & - & 70.7 & \cellcolor{blue!30}75.2 & - & \cellcolor{blue!30}76.0 & \cellcolor{blue!10}73.1 & - & \cellcolor{blue!40}78.7 & \cellcolor{blue!30}75.0 \\
\midrule



Mistral$_{123B}$ & 61.9 & \cellcolor{blue!10}70.1 & \cellcolor{blue!20}73.9 & 62.2 & 69.6 & \cellcolor{blue!10}70.8 & 63.3 & 71.6 & \cellcolor{blue!40}77.8 & 63.5 & 71.5 & \cellcolor{blue!20}74.7 & 62.6 & 69.6 & \cellcolor{blue!20}73.6\\  
\textit{   +CoT}  & - & 67.8 & \cellcolor{blue!10}72.4 & - & 62.3 & 70.1 &- & 69.6 & \cellcolor{blue!30}74.6 & - & 65.5 & \cellcolor{blue!10}72.0 & -& 68.1 & \cellcolor{blue!10}71.2 \\
\midrule

LLaMA 3.1$_{70B}$ & 68.9 & \cellcolor{blue!10}72.9 & \cellcolor{blue!20}73.7 & 69.9 & \cellcolor{blue!10}70.3 & \cellcolor{blue!30}73.9 & 70.1 & \cellcolor{blue!10}71.7 & \cellcolor{blue!20}73.3 & 67.8 & 70.9 & \cellcolor{blue!20}75.3 & 70.5 & \cellcolor{blue!20}73.0 & 68.7 \\  
\textit{+CoT}  & - & 66.1 & 69.7 & - & 64.3 & 67.3 & - & 69.7 & \cellcolor{blue!20}72.8 & - & 64.4 & 68.5 & - & 68.7 & 70.2 \\ 

\midrule

LLaMA 3.1$_{8B}$ & 51.9 & 60.5 & 55.9 & 51.2 & 59.1 & 51.1 & 51.7 & 59.1 & 58.4 & 57.5 & 56.7 & 53.7 & 52.6 & 62.6 & 57.3 \\   
\textit{   +CoT}  & - & 48.4 & 54.6 & - & 49.9 & 55.1 & - & 47.0 & 56.1 & - & 49.7 & 51.6 & - & 46.7 & 60.8 \\
\midrule

LLaMA 3.1$_{8B}$ (Base) & 49.6 & 50.5 & 50.2 & 50.4 & 51.7 & 50.0 & 49.3 & 53.6 & 50.3 & 49.6 & 50.8 & 50.1 & 47.6 & 48.2 & 48.8 \\
\textit{   +CoT}  & -& 52.3 & 50.8 & - & 49.9 & 53.2 & - & 52.6 & 50.9 & - & 53.0 & 51.9 & - & 51.7 & 52.4 \\
\midrule

LLaMA 3.2$_{3B}$ & 52.2 & 53.5 & 51.0 & 51.7 & 52.3 & 47.2 & 55.9 & 55.1 & 51.0 & 54.7 & 51.2 & 51.3 & 53.5 & 52.6 & 51.7 \\
\textit{   +CoT} & - & 53.8 & 57.2 & - & 53.7 & 57.1 & - & 53.0 & 55.9 & -& 53.6 & 59.5 & - & 49.9 & 56.5 \\
\midrule

LLaMA 3.2$_{3B}$ (Base) & 49.8 & 49.5 & 49.3 & 50.0 & 57.6 & 47.9 & 47.7 & 48.7 & 48.5 & 49.8 & 54.9 & 49.8 & 49.3 & 45.8 & 45.4\\
\textit{+CoT}  & - & 48.5 & 47.5 & -& 48.4 & 47.5 & - & 51.1 & 50.3 & - & 49.4 & 49.3 & - & 47.8 & 48.6\\
\midrule

LLaMA 3.2$_{1B}$ & 49.9 & 47.7 & 48.9 & 48.7 & 48.6 & 51.1 & 51.6 & 48.8 & 51.2 & 49.6 & 48.4 & 48.7 & 49.8 & 46.7 & 52.3 \\
\textit{   +CoT}  & - & 49.9 & 50.1 & -& 45.3 & 52.1 & -& 45.7 & 48.7 & - & 46.7 & 50.2 & - & 47.6 & 49.0 \\
\midrule

LLaMA 3.2$_{1B}$ (Base) & 47.5 & 52.4 & 48.7 & 45.4 & 52.3 & 49.4 & 49.8 & 49.6 & 51.2 & 48.1 & 50.9 & 46.1 & 48.8 & 48.7 & 48.1 \\
\textit{+CoT} & - & 51.0 & 51.0 & - & 50.5 & 38.5 &- & 51.1 & 51.7 & - & 50.5 & 50.5 & - & 52.2 & 49.3 \\
\midrule

Flan-T5$_{780M}$ & 57.2 & 56.3 & 55.5 & 55.3 & 54.6 & 53.5 & 57.4 & 58.3 & 57.8 & 55.3 & 54.5 & 52.4 & 60.4 & 59.0 & 59.9 \\
\textit{   +CoT} & - & 58.2 & 55.9 & - & 56.0 & 54.3 & - & 59.6 & 57.8 & - & 55.3 & 54.2 & - & 61.4 & 59.0\\
        
\bottomrule
\end{tabular}}

\caption{\corpus\xspace experimental results (accuracy) under zero-shot (ZS), one-shot (1S), and few-shot (FS) settings for standard and CoT prompting. All results are averaged on three runs. A gradient-based color scheme highlights higher accuracy scores, with darker shades indicating better performance.}
\label{table:results}
\end{table*}

\paragraph{Citations.}
Legal texts frequently include citations (23.68\% of sentences), anchoring arguments through references to past court decisions and statutes, often encoding past events.

\paragraph{Broader range of verb tenses and speculative events.}
In \textsc{Tracie}, most events appear in the simple past (87.62\%), while \corpus\space shows more tense variation, with 24.24\% in the present simple and 5.57\% in the present perfect
(Appendix \ref{app: chart_tenses}). Additionally, speculative events are more common in \corpus\space (9.83\%, +4.1\%). Tense variation affects temporal reasoning, as some tenses signal rules and precedents. For instance, the present perfect marks past events with ongoing relevance (e.g., \textit{The plaintiff has submitted an appeal}).


\section{Benchmark Methodology}\label{sec:methodo}
We conduct evaluations using two different prompt-based approaches: standard prompting and chain-of-thought (CoT) prompting. We perform experiments in zero-shot (ZS), one-shot (1S), and few-shot (FS) settings (3 examples). Prompts and examples can be found in Appendix \ref{app: prompts_eval}. 

\paragraph{Standard prompting.} In a ZS setting, the model is prompted to answer a query containing the triplet \textit{\{event A, TR, event B\}} based on a context paragraph. In 1S and FS settings, models are provided with one and three examples (context paragraphs and question-answer pairs), respectively.

\paragraph{CoT prompting.} Instructions for CoT prompts are the same as those for standard prompting. CoT prompts guide the LM to reason through steps before giving a final answer. We do not use CoT in a ZS setting, as we consider examples necessary. We manually annotate up to three examples for step-by-step reasoning in 1S and FS settings. 

\paragraph{Evaluation.} We evaluate our benchmark using accuracy. The evaluation compares long and short inputs to reflect a key characteristic of legal language (sentence and context length), which captures finer linguistic complexities such as conditional clauses, subordination, and event density. There are 230 paragraphs with fewer than 150 tokens (short) and 282 paragraphs with more than 150 tokens (long). Inspired by \textsc{Tracie} and based on our linguistic analysis, we address the importance of implicit event reasoning in legal NLP. We evaluate the influence of implicit events considering 210 explicit-implicit pairs, compared to 288 explicit-explicit pairs (and 14 implicit-implicit pairs).

\paragraph{Models and experimental setup.} We evaluate a range of popular LMs, including both open-source and proprietary models (Appendix \ref{app: models}), spanning from smaller models (LLaMA 3.2$_{1B}$, Flan-T5$_{780M}$) to larger ones (GPT-4). Proprietary models are accessed through their APIs, non-proprietary models are accessed through HuggingFace and used in lower precision formats (16-bit floating-point and 4-bit quantization).

\section{Experimental Results} \label{sec:results}

\begin{tcolorbox}[colback=blue!2!white,colframe=blue!30!black, boxrule=0.4pt, arc=2pt]
  \footnotesize GPT-4o, GPT-4 Turbo outperform Mistral, LLaMA, and FLAN models across all evaluated prompts and settings.
\end{tcolorbox}

Experimental results show that LLMs perform better on \corpus{} than on \textsc{Tracie}, suggesting their ability to tackle some linguistic complexities of legal language. However, \S\ref{sec:errors_analysis} will show that these complexities still limit performance.

Across the five studied subsets, GPT-4 Turbo achieves the highest accuracy on \corpus{} in four cases (up to 80.8\% accuracy for explicit-implicit pairs of events), including three in an FS setting and one in a 1S CoT setting (all data). The best accuracy for the short-context subset is obtained by Mistral$_{123B}$ in a FS setting with 77.8\%. Overall, CoT appears among the top three results only three times across all categories (all data, long context, and explicit-implicit), suggesting its limited impact on performance in this task.

Compared to GPT-4o, Mistral$_{123B}$ exhibits lower accuracy across all prompts, with performance differences ranging from 0.2\% to 8\%. LLaMA 3.1$_{70B}$ also shows reduced accuracy, with a performance gap between 1\% and 3.7\%. Flan-T5$_{780M}$ demonstrates a significant decline, with accuracy differences ranging from 10.4\% to 21.9\%. When considering the other LLaMA models collectively, their accuracy is lower by 14.4\% to 28.7\%.

Compared to GPT-4 Turbo, Mistral$_{123B}$ shows lower accuracy across all prompts, with differences ranging from 1.9\% to 9.8\%. LLaMA 3.1$_{70B}$ exhibits a decline in accuracy ranging from 1.3\% to 11.5\%, while Flan-T5$_{780M}$ shows a more significant drop, with differences between 13\% and 21.7\%. The other LLaMA models collectively show accuracy reductions ranging from 14.6\% to 29.2\%.

\subsection{Chain-of-Thought in Event Ordering}

\begin{figure}[H]
\includegraphics[width=0.8\columnwidth]{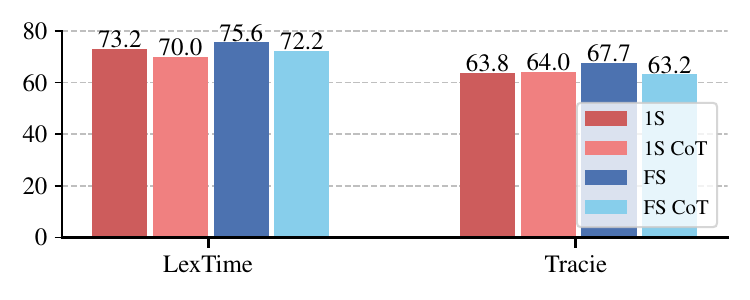} 
\centering
   \caption{Performance gap (accuracy) with and without CoT
prompting. Results are averaged from GPT-4, GPT-4 Turbo, LLaMA 3.1$_{70B}$ and Mistral$_{123B}$.}
   \label{figure: cot_chart}
\end{figure}

\begin{tcolorbox}[colback=blue!2!white,colframe=blue!30!black, boxrule=0.4pt, arc=2pt]
  \footnotesize CoT prompting does not improve accuracy for event ordering.
\end{tcolorbox}

Prior research has demonstrated that CoT prompting can improve a model's reasoning skills \cite{10.5555/3600270.3602070}. As illustrated in Fig. \ref{figure: cot_chart}, CoT prompting does not improve performance for \corpus\space or \textsc{Tracie}. The best results are achieved using an FS approach with a standard prompt (75.6\%). Employing CoT in FS reduces performance by 3.4\%. In a 1S setting, we also observe a decrease in accuracy of 3.2\%. Similar results are seen with the \textsc{Tracie} dataset. Table \ref{table:results} shows that these findings hold true regardless of context length and implicit events. This indicates that incorporating elicited reasoning steps did not improve the model's ability to retrieve temporal relationships, possibly because of the increased prompt length when including multiple examples. Longer prompts may lead to attention dilution — as the prompt length increases, attention is distributed across more tokens, making it harder to distinguish key reasoning steps from less relevant information, as discussed in \citet{shi2023largelanguagemodelseasily}.

\subsection{Impact of Context Length}

\begin{tcolorbox}[colback=blue!2!white,colframe=blue!30!black, boxrule=0.4pt, arc=2pt]
  \footnotesize Larger LMs benefit from longer contexts, smaller LMs perform better with shorter context windows.
\end{tcolorbox}

GPT-4 Turbo achieves the highest accuracy in long-context settings (79.9\%), while Mistral$_{123B}$ performs best in short-context settings (77.8\%), both with a FS prompting setup.

For GPT-4o, GPT-4 Turbo, and LLaMA 3.1$_{70B}$, increasing the input context length leads to improved performance, with GPT-4 Turbo achieving up to a 3.5\% gain on a FS setting. In contrast, all other evaluated models exhibit higher accuracy with shorter context lengths (below 150 tokens), with Mistral$_{123B}$ demonstrating a 7\% improvement and LLaMA 3.1$_{8B}$ achieving up to a 7.3\% increase. This observation suggests that larger models, such as GPT-4 Turbo, leverage the additional contextual information to capture temporal structures and event relationships, ultimately leading to more accurate predictions. In settings other than FS, we observe that shorter context lengths also enhance performance. Specifically, GPT-4 Turbo achieves a 3.9\% improvement with shorter context windows in the CoT few-shot setting, and LLaMA 3.1$_{70B}$ demonstrates a performance gain of 5.5\%.

\subsection{Impact of Implicit Events}

\begin{tcolorbox}[colback=blue!2!white,colframe=blue!30!black, boxrule=0.4pt, arc=2pt]
  \footnotesize GPT-4 Turbo achieves the highest overall accuracy at 80.8\% for explicit-implicit pairs, outperforming the best result on the full dataset by +3.2\%.
\end{tcolorbox}
 
GPT-4 Turbo achieves the highest accuracy across both subsets (78.8\% and 80.8\%, FS). Several hypotheses may explain why queries containing one explicit and one implicit event achieve higher scores. One possible explanation is due to the dataset structure: implicit events often appear near explicit ones, which may make the temporal relation between the two events more constrained and easier for the model to infer. By contrast, two explicit events can occur in distant parts of the text or be separated by complex subordinate clauses and repeated paraphrases.

GPT-4o, GPT-4 Turbo, and Flan-T5$_{780M}$ yield better accuracy on explicit-implicit event pairs, improving by +2.2\%, +2.6\%, and +7.5\%, respectively, in their FS settings. This suggests that these models are capable of inferring missing temporal information. In contrast, Mistral$_{123B}$ and LLaMA 3.1$_{70B}$ perform better on explicit-explicit pairs, achieving improvements of +1.1\% and +6.6\% in the FS settings. This suggests that these models rely on direct lexical and temporal cues.

When considering all LLaMA models except LLaMA 3.1$_{70B}$, base models exhibit a larger performance improvement on explicit-explicit pairs (+4.7\%) compared to explicit-implicit pairs, while we observe a slight decrease with aligned models (-0.4\%) — on average across all prompt types. Overall, for the aligned LLaMA models (excluding 70B), the average accuracy for explicit-implicit pairs is 52.6\%, compared to 45.6\% for base models. This suggests that alignment enhances implicit reasoning capabilities.

\subsection{Comparison with \textsc{Tracie}}
\begin{table}[t]
\centering
\resizebox{0.80\columnwidth}{!}{
\begin{tabular}{l|cccc}
\toprule
  & ZS  & 1S  &FS\\ 
\midrule

GPT-4o&  67.0  (\textcolor{ red}{\(\downarrow\)}   2.9) &67.4  (\textcolor{ red}{\(\downarrow\)}   7.5)
&66.9  (\textcolor{ red}{\(\downarrow\)}   10.5)\\
\textit{   +CoT}& - &61.8 (\textcolor{ red}{\(\downarrow\)}   6.8)&64.0  (\textcolor{ red}{\(\downarrow\)}   8.6)\\
\midrule

GPT-4 Turbo & 70.8 (\textcolor{ green}{\(\uparrow\)}   0.6)&66.5 (\textcolor{ red}{\(\downarrow\)}   8.6)&\textbf{73.1} (\textcolor{ red}{\(\downarrow\)}   4.1) \\
\textit{   +CoT}&-&\textbf{73.6} (\textcolor{ red}{\(\downarrow\)}   4.0)&65.9  (\textcolor{ red}{\(\downarrow\)}   8.4)\\
\midrule



Mistral$_{123B}$ & 63.9 (\textcolor{ green}{\(\uparrow\)}   2)&61.1 (\textcolor{ red}{\(\downarrow\)}   9)&64.6 (\textcolor{ red}{\(\downarrow\)}   9.3)\\
\textit{   +CoT} &-&58.4 (\textcolor{ red}{\(\downarrow\)}   9.4)&\textbf{72.4} (\textcolor{blue}{\(=\)} 0.0)\\
\midrule

LLaMA 3.1$_{70B}$& 62.3 (\textcolor{ red}{\(\downarrow\)}  6.6 )&63.0 (\textcolor{ red}{\(\downarrow\)} 9.9  )&66.2 (\textcolor{ red}{\(\downarrow\)} 7.5  )\\
\textit{   +CoT} &-&62.0 (\textcolor{ red}{\(\downarrow\)} 4.1  )&65.5 (\textcolor{ red}{\(\downarrow\)} 4.2  )\\

\midrule

LLaMA 3.1$_{8B}$ &53.5 (\textcolor{ green}{\(\uparrow\)}   1.6)&54.9 (\textcolor{ red}{\(\downarrow\)}   5.6)&52.7 (\textcolor{ red}{\(\downarrow\)}   3.2)\\
\textit{   +CoT} &-&54.6 (\textcolor{ green}{\(\uparrow\)}   6.2)&56.4 (\textcolor{ green}{\(\uparrow\)}   1.8)\\
\midrule

LLaMA 3.1$_{8B}$ (Base) &50.0 (\textcolor{ green}{\(\uparrow\)}   0.4)&48.9 (\textcolor{ red}{\(\downarrow\)}   1.6)&50.1  (\textcolor{ red}{\(\downarrow\)}   0.1)\\
\textit{   +CoT} &-&49.5 (\textcolor{ red}{\(\downarrow\)}   2.8)&51.8 (\textcolor{ green}{\(\uparrow\)}   1.0)\\
\midrule

LLaMA 3.2$_{3B}$ &53.4 (\textcolor{ green}{\(\uparrow\)}   1.2)&53.5 (\textcolor{blue}{\(=\)} 0.0)&53.5 (\textcolor{ green}{\(\uparrow\)}   2.5)\\
\textit{   +CoT} &-&51.6 (\textcolor{ red}{\(\downarrow\)}   2.2)&51.2 (\textcolor{ red}{\(\downarrow\)}   6.0)\\
\midrule

LLaMA 3.2$_{3B}$ (Base) 
& 54.2 (\textcolor{ green}{\(\uparrow\)} 4.4  )

&50.9 (\textcolor{ green}{\(\uparrow\)} 1.4  )

&50.8 (\textcolor{ green}{\(\uparrow\)} 1.5  )\\
\textit{   +CoT} &-
&50.8 (\textcolor{ green}{\(\uparrow\)} 2.3  )

&48.2 (\textcolor{ green}{\(\uparrow\)} 0.7  )\\
\midrule

LLaMA 3.2$_{1B}$ &50.3 (\textcolor{ green}{\(\uparrow\)}   0.9)&50.2 (\textcolor{ green}{\(\uparrow\)}   2.5)&51.1 (\textcolor{ green}{\(\uparrow\)}   2.2)\\
\textit{   +CoT} &-&49.1 (\textcolor{ red}{\(\downarrow\)}   0.8)&49.2 (\textcolor{ red}{\(\downarrow\)}   0.9)\\
\midrule

LLaMA 3.2$_{1B}$ (Base) & 49.4 (\textcolor{ green}{\(\uparrow\)} 1.9  )&51.1 (\textcolor{ red}{\(\downarrow\)} 1.3  )&51.7 (\textcolor{ green}{\(\uparrow\)}  3 )\\
\textit{   +CoT} &-&50.9 (\textcolor{ red}{\(\downarrow\)} 0.1 )&45.6 (\textcolor{ red}{\(\downarrow\)} 5.4 )\\
\midrule

Flan-T5$_{780M}$ &53.9 (\textcolor{ red}{\(\downarrow\)}   3.3)&53.9 (\textcolor{ red}{\(\downarrow\)}   2.4)&54.3 (\textcolor{ red}{\(\downarrow\)}   1.2)\\
\textit{   +CoT} &-&54.6 (\textcolor{ red}{\(\downarrow\)}   3.6)&54.9 (\textcolor{ red}{\(\downarrow\)}   1.0)\\
\bottomrule
\end{tabular}}
\caption{ \textsc{tracie} experimental results (accuracy) under zero-shot (ZS), one-shot (1S), and few-shot (FS) settings for standard prompting and CoT. Top 3 results are in bold. Values in parentheses indicate the absolute difference between \corpus\space and \textsc{Tracie}. \textcolor{ green}{\(\uparrow\)} signifies that accuracy is higher on \textsc{Tracie} compared to \corpus. \textcolor{ red}{\(\downarrow\)} indicates that accuracy is lower on \textsc{Tracie}. }
\label{table:tracie_results}
\vspace{-6pt}
\end{table}

\begin{tcolorbox}[colback=blue!2!white,colframe=blue!30!black, boxrule=0.4pt, arc=2pt]
  \footnotesize The highest accuracy achieved on \textsc{Tracie} at 73.6\% (GPT-4 Turbo) is 4\% lower than the highest accuracy on \corpus.
\end{tcolorbox}

\begin{figure*}[ht]
\includegraphics[width=0.9\textwidth]{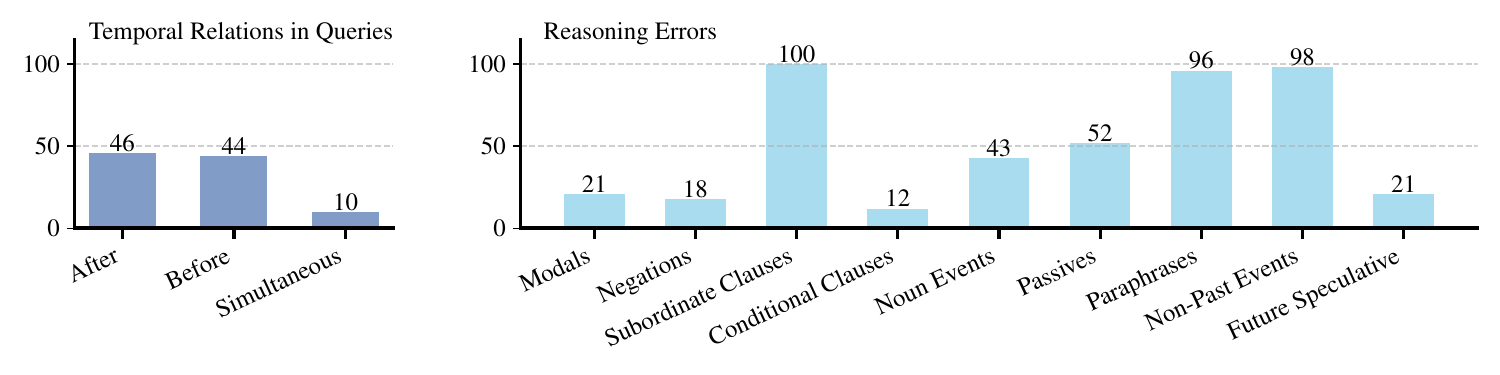} 
\centering
   \caption{Error analysis for \corpus\space on 100 misclassified samples from the few-shot setting across 
GPT-4, GPT-4 Turbo and LLaMA 3.1$_{70B}$. The left chart shows the TR distribution in the queries, and the right plot categorizes errors by counting features found in the context paragraphs. \textbf{Takeaway:} Reasoning errors are disproportionately associated with the linguistic specificities of legal language identified in \S\ref{sec:linguistic_analysis}.}
   \label{figure: errors}
\end{figure*}

GPT-4o, LLaMA 3.1$_{70B}$, and FLAN-T5 consistently perform better on \corpus\space across all prompt settings, with improvements of up to 10.5\% (GPT-4o, FS). GPT-4 Turbo and Mistral$_{123B}$ generally perform better on \corpus\space, except in the ZS setting. For the smaller LLaMA models (all except 70B), \corpus\space outperforms \textsc{Tracie} by up to 6\% in the CoT FS setting (LLaMA 3.2$_{3B}$). However, in the 1S CoT, \textsc{Tracie} surpasses \corpus\space by 6.2\%, indicating that task performance varies depending on the prompting strategy.

The accuracy difference suggests distinct challenges between legal and narrative event ordering. One possible explanation for the lower accuracy on \textsc{Tracie} may be longer contexts in \corpus\space (Fig. \ref{figure: boxplot}), which provides more contextual information. A second hypothesis relates to less explicit and potentially more challenging language for temporal reasoning (Fig. \ref{figure: language_analysis}) in \textsc{Tracie}.

\section{Errors Analysis} \label{sec:errors_analysis}

Building on the linguistic analysis in \S\ref{sec:linguistic_analysis}, we manually analyze 100 erroneous predictions by GPT-4, GPT-4 Turbo, and LLaMA 3.1$_{70B}$. Visualization of errors is shown in Fig. \ref{figure: errors}.

\paragraph{TR in queries.} We analyze the distribution of errors by categorizing according to the three defined temporal relationship types. Errors are evenly distributed between \textit{After} and \textit{Before} TRs, with fewer \textit{Simultaneous} errors. This pattern reflects the initial dataset composition (\S \ref{sec:dataset}), indicating no bias toward any specific temporal relation type.

\paragraph{Reasoning errors.}
We find that reasoning errors are disproportionately associated with the linguistic specificities of legal language identified earlier (\S\ref{sec:linguistic_analysis}). Compared to the overall dataset (Fig. \ref{figure: language_analysis}), errors are associated with a +89\% increase in paraphrases, +84\% in subordinate clauses, +63\% in events not in the past simple tense, +24\% in passive voice, +12\% in negations and future speculative constructions, +6\% in modality markers, and +2\% in conditional clauses. This suggests that these linguistic phenomena contribute to the difficulty of event ordering in legal texts.

We analyze the characteristics of context paragraphs that led to errors in event ordering. Several features appear in more than half of the analyzed error cases: subordinate clauses, non-past events, paraphrases, and passive constructions.

(1) Subordinate clauses embed events in hierarchical structures, complicating dependency relations between events. For example: \begin{quote}
\textit{The applicant Mrs.~X \underline{learned} of the misuse of union funds only in March~2021, 
when Mr.~Y, the vice president at the time, \underline{obtained} copies of her weekly expense reports from~2015.}
\end{quote} Here, the event queried (\textit{obtained copies of her weekly expense reports}) occurs within a subordinate clause introduced by \textit{when}, rather than in the main clause. If we wanted to order the two main events in this sentence (event A as \textit{Mrs X learned of the misuse} and Event B as \textit{Mr Y obtains copies of expense reports}), we note that event B in the subordinate clause happens before event A, but occurs later in the sentence. This creates a discrepancy between the order of the words and the actual order of the events.  

(2) Events not expressed in the past simple tense are +63\% more frequent in \corpus{} than in \textsc{Tracie}. Tense variation increases temporal complexity, compared to texts where most events occur in the same tense. This is particularly the case when verb tense signals specific legal functions, such as rules, obligations, or precedents -- patterns that are less common in generic narratives. Consequently, we hypothesize that models trained primarily on non-legal corpora struggle to interpret these temporal cues.

(3) Legal texts frequently employ paraphrases to increase clarity or restate obligations, which can lead to multiple mentions of the same event.

(4) Passives often obscure the agent or temporal anchor of an event, making it harder for models to determine when an action occurred and who performed it.

While \S\ref{sec:linguistic_analysis} highlighted nominalizations as a key mechanism for presenting events in legal texts, \corpus{} does not include nouns as event triggers. Nevertheless, noun events appear in the context paragraphs, which explains their presence in the error analysis Fig. \ref{figure: errors}.

Overall, the error patterns suggest that the linguistic phenomena most characteristic of legal discourse are precisely those that hinder event ordering accuracy with language models. This reinforces the claim that improving legal temporal reasoning requires domain-sensitive modeling.


\section{Related Work} \label{sec:related_work}

\paragraph{Benchmarking LLMs.} 
Previous datasets and benchmarks \cite{10.1007/978-3-031-28244-7_28, timeqa_neurips2024}
study the performance of LLMs on temporal reasoning, showing that those tasks pose a challenge and that the limitations of LLMs in temporal reasoning are not well understood. They show that LMs generally struggle with temporal reasoning, as evidenced by \citet{feng-etal-2023-generic}, who demonstrate that LLMs often take random guesses. \citet{jain-etal-2023-language-models} show that LLMs perform better on event duration tasks than on event ordering tasks and that reasoning over longer contexts is more difficult, two aspects we study in \corpus.

\citet{qiu-etal-2024-large} indicate that LLMs lag behind human performance and small-scale but specialized LMs. The performance gap is only marginally reduced by introducing in-context learning, fine-tuning, and CoT prompting. Their analysis shows that current LLMs struggle with having a consistent temporal model due to insufficient exposure during pretraining. We investigate the effectiveness of CoT prompting within the legal domain.

\paragraph{Temporal reasoning and understanding.}
Grounding events in a temporal context has a notable history in NLP. \citet{pustejovsky2005temporal} demonstrated that temporal grounding affects question-answering model quality and defined an event as a situation that occurs or holds true at a specific time point or interval. Recent temporal reasoning benchmarks advanced our understanding of LMs' capabilities \cite{timeqa_neurips2024, 10.1007/978-3-031-28244-7_28}. However, existing datasets primarily focus on general or narrative texts \cite{zhou-etal-2021-temporal}. Temporal reasoning consists of subtasks such as event frequency, duration, or ordering \cite{zhou-etal-2019-mctaco, jain-etal-2023-language-models}. Event ordering is particularly important for constructing timelines in the legal domain. Despite its significance, no prior work has systematically evaluated LMs’ ability to perform temporal event ordering in legal texts.

Existing datasets include several approaches: natural language inference and relation extraction \cite{thukral-etal-2021-probing, 10.1007/978-3-031-28244-7_28, zhou-etal-2021-temporal}, question-answering (QA), such as and \textsc{MC-TACO} \cite{zhou-etal-2019-mctaco}, \textsc{TimeDial} for dialogue settings \cite{qin-etal-2021-timedial}, \textsc{TempReason} \cite{tan-etal-2023-towards}, and \textsc{TimeQA} \cite{timeqa_neurips2024}. Our work is inspired by those methods, framing the task as a simplified QA format with binary labels.

\section{Conclusion}
We present \corpus, the first benchmark for evaluating temporal event ordering in legal language, and systematically evaluate LLMs' ability to reason over legal temporal structures. Our findings show that while LLM performance on \corpus{} is superior to their performance on a narrative text benchmark, the complexities of legal discourse, such as paraphrasing, constrain accuracy. We show that longer input contexts and implicit-explicit event pairs improve model accuracy, offering practical insights for optimizing LLM-based legal NLP applications. Experimental results suggest that further performance improvements could rely on exploiting identified errors and linguistic structure to align models with specialized domain language.

\section{Limitations}
Our study has several limitations. First, we do not provide a comprehensive evaluation of ambiguous cases, as these are classified and resolved during the annotation stage. Second, the temporal relations scheme is limited to three categories, whereas a more granular approach would be necessary to handle more scenarios and would capture richer temporal distinctions.

The generalizability of our findings is constrained by the specificity of the area of the law and by the size of \corpus\space (512 samples). While our benchmark represents an initial step in evaluating temporal reasoning in legal texts, it primarily focuses on event ordering. Future research should extend this evaluation to include other temporal reasoning tasks, such as event frequency and duration analysis, to achieve a more comprehensive understanding.

\section*{Acknowledgments}
The work was completed while Claire Barale was a Bloomberg Data Science Ph.D. Fellow. Experiments have made use of the resources provided by the Edinburgh Compute and Data Facility (ECDF) and the Edinburgh International Data Facility (EIDF).

\bibliography{custom}
\appendix

\section*{Appendix}
\label{sec:appendix} 

\section{Specific Features of Legal Events} \label{app: linguistic_analysis}
\subsection{\textsc{Tracie} Event Triggers} \label{app: tracie_triggers}

\begin{figure}[H]
\includegraphics[width=0.9\columnwidth]{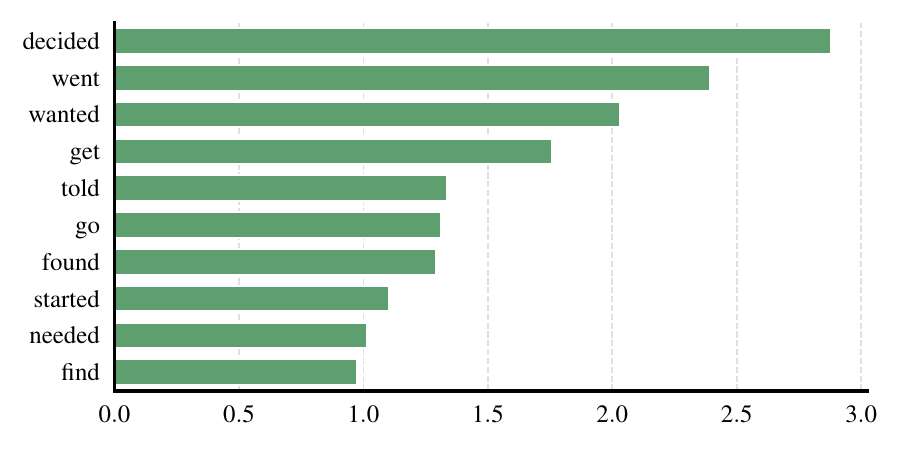} 
\centering
   \caption{Top-10 event triggers in \textsc{Tracie} by frequency (\%).}
   \label{figure: tracieevent_triggers}
\end{figure}

\subsection{Keywords for Temporal Markers Extraction} \label{app: temp_keywords}
\begin{itemize}[leftmargin=*]
    \item \textbf{General time:} \textit{now, then, before, after, later, soon, earlier, previously, recently, already, immediately, eventually, finally, formerly, subsequently, yet, ever, until, till, since, while, when, whenever, as soon as, as long as, afterwards, meanwhile, in the meantime, up to now, up until now.}
    
    \item \textbf{Frequency Adverbs:} \textit{always, usually, often, sometimes, occasionally, rarely, seldom, never, frequently, constantly, continuously, periodically, infrequently, repeatedly}
\end{itemize}
Adverbs, conjunctions or prepositions that don't appear in Fig. \ref{figure: adverbs_frequency} were not found in the datasets.

\subsection{Frequency of Temporal Cues} \label{app: chart_temp_adverbs}
\begin{figure*}[t]
\includegraphics[width=\textwidth]{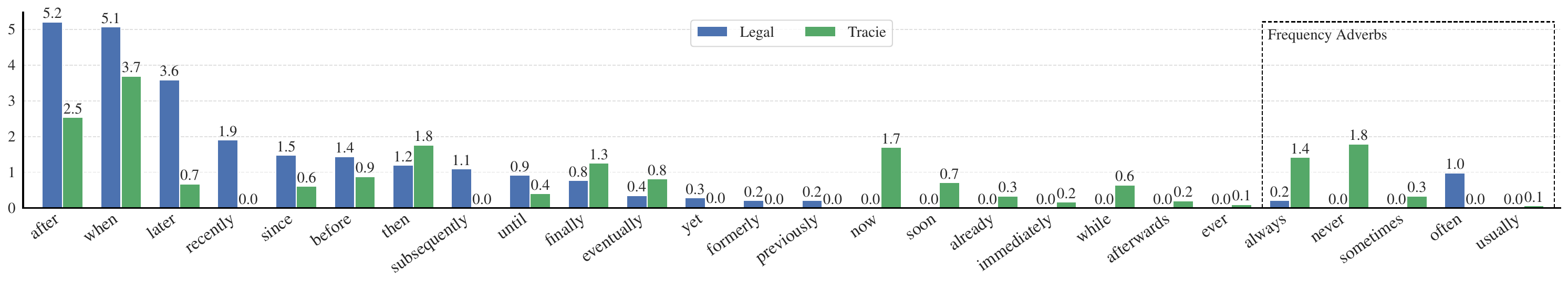} 
\centering
   \caption{Comparison of the frequency of temporal adverbs in \corpus\space (blue) and \textsc{Tracie} (green). Values (y-axis) represent the percentage of sentences that contain each adverb, indicating its frequency in the text.}
   \label{figure: adverbs_frequency}
\end{figure*}

See Fig. \ref{figure: adverbs_frequency}.

\begin{figure*}[t]
\includegraphics[width=\textwidth]{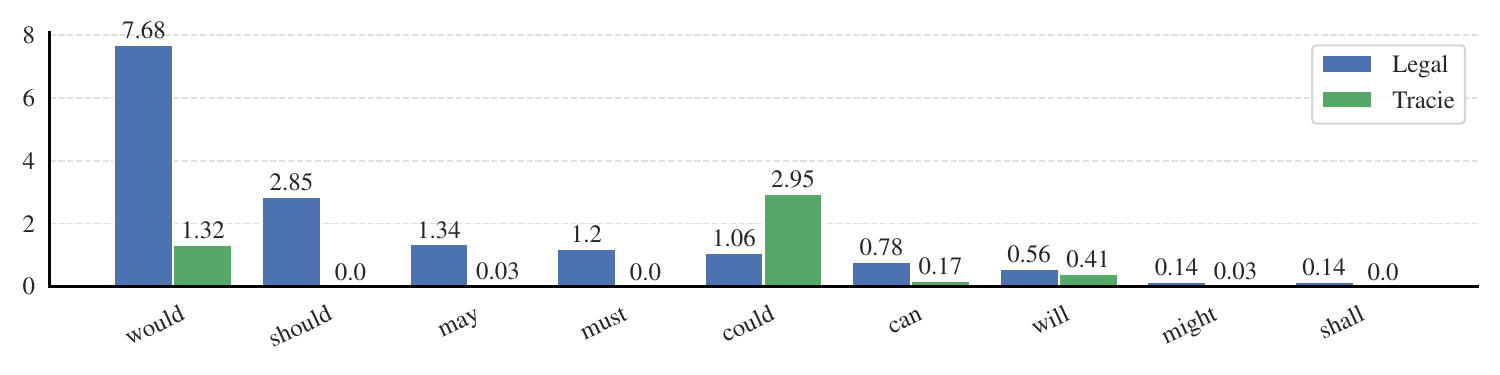} 
\centering
   \caption{Comparison of the frequency of the modal verbs in \corpus\space (blue) and \textsc{Tracie} (green). Values (y-axis) represent the percentage of sentences that contain each modal, indicating its frequency in the text.}
   \label{figure: modal_frequency}
\end{figure*}

\subsection{Frequency of Modal Verbs}\label{app: chart_modals}
See Fig. \ref{figure: modal_frequency}.

\subsection{Frequency of Verb Tenses}\label{app: chart_tenses}
\begin{figure}[H]
\includegraphics[width=\columnwidth]{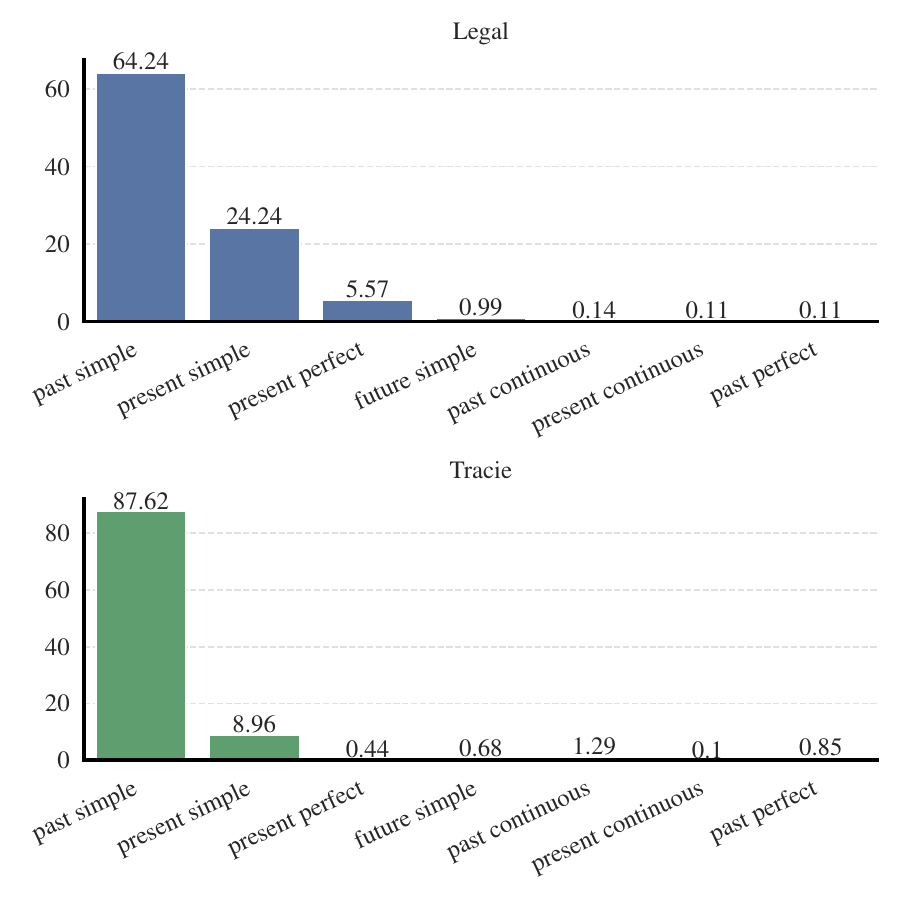} 
\centering
   \caption{Distribution of the tenses of the verbs in \corpus\space and \textsc{Tracie}. The y-axis is the frequency, in \%. The y-axis scales differ between the charts for readability.}
   \label{figure: verbs}
\end{figure}

Fig. \ref{figure: verbs} shows the frequency of the tenses found in the context paragraph, for both datasets. 

\section{Models}\label{app: models}

\subsection{Impact of Model Size}\label{app: modelsize}
\begin{figure}[H]
\includegraphics[width=\columnwidth]{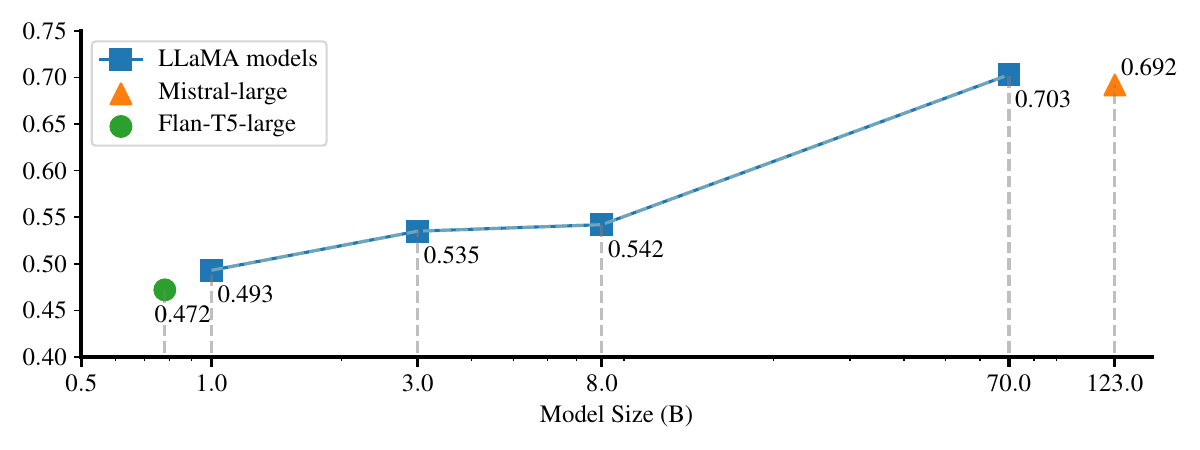} 
\centering
   \caption{Scaling effect of model size and overall temporal reasoning performance. The y-axis shows the average accuracy across all prompt types for each model. The x-axis (model size) is shown in the log scale. \textbf{Takeaway:} Results show a log-linearity between parameter size and performance.}
   \label{figure: scaling_cahrt}
\end{figure}

\subsection{Models Used}
\paragraph{GPT-4o and GPT-4-turbo.} 
GPT-4-turbo refers to \textit{gpt-4-turbo-2024-04-09}, and GPT-4o refers to \textit{gpt-4o-2024-08-06}. GPT-4o provides faster inference compared to turbo \cite{openai2023gpt}. 

\paragraph{Llama 3.1 and Llama 3.2.}
We use the HuggingFace implementation of \textit{Llama-3.1-8B, Llama-3.1-8B-Instruct, Llama-3.1-70B, Llama-3.1-70B-Instruct, Llama-3.2-1B, Llama-3.2-1B-Instruct, Llama-3.2-3B, Llama-3.2-3B-Instruct}. Llama 3.1 and 3.2 fine-tuned versions use supervised fine-tuning (SFT) and reinforcement learning from human feedback (RLHF), Rejection Sampling (RS), and Direct Preference Optimization (DPO, \cite{dubey2024llama}). 

\paragraph{Mistral.} We use the large version, 123B parameters, \textit{Mistral-Large-Instruct-2407}. 

\paragraph{FLAN-T5-large.} We use the large version of FLAN-T5 \cite{chung2024scaling}. FLAN-T5 is built on top of T5 \cite{raffel2020exploring} through instruction fine-tuning.

\subsection{GPU Resources}\label{app: gpus}
This work was made possible by the use of: Nvidia Tesla V100-SXM2, Nvidia Tesla K80, Nvidia A100 40GB and 80GB, and Nvidia RTX A6000 40GB. 

\section{Dataset Construction}\label{app: dataset_example}
\subsection{Examples from \corpus}\label{B2_examples}
\begin{figure*}[h!]
\centering
\includegraphics[width=\textwidth]{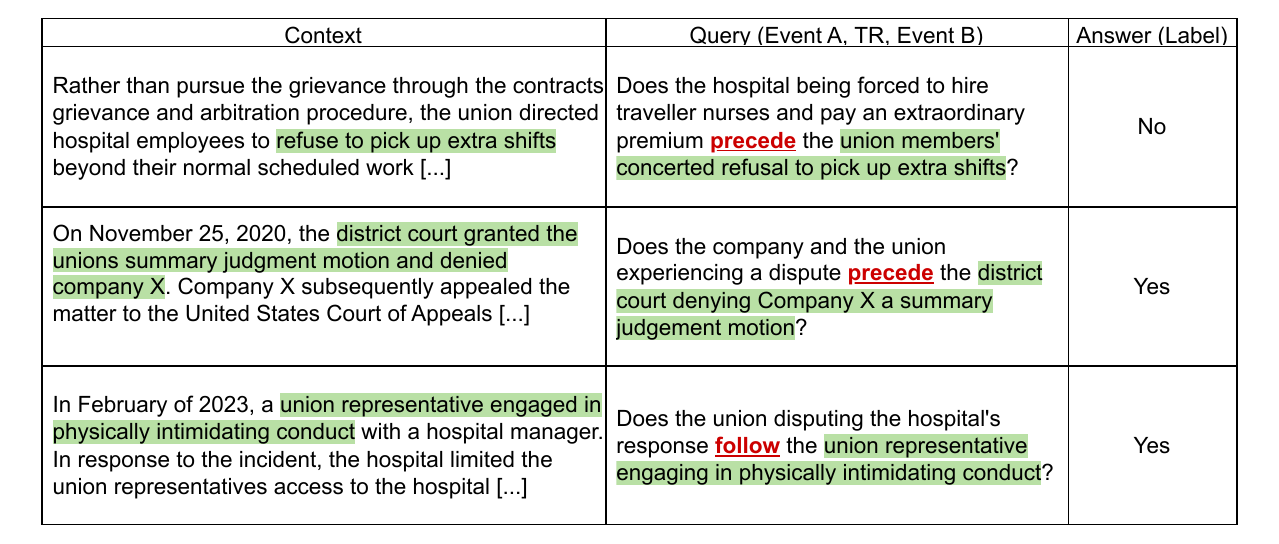} 
  \caption{Examples from \corpus, showing three instances composed of a context paragraph, a query, and a binary label. Event B (explicit) is highlighted in green in the context and the query. Event A from the query occurs later in the context, beyond the visible portion.}
   \label{figure: examples}
\end{figure*}

See Fig. \ref{figure: examples}.

\subsection{Prompts for Dataset Generation}\label{app: prompts_dataset_generation}
See Table \ref{table:datagen}.

\begin{table*}[h!]
\centering
 
\resizebox{\textwidth}{!}{
\begin{tabular}{lp{15cm}}

\textbf{Description} & \textbf{Prompt} \\ 
\midrule
\cellcolor{gray!20} Prompt 1: Get paragraphs & Your task is to find if \{context\_paragraph\} contains events. The events should be verbs. If it does not contain relevant events, just answer with "None" (no explanation). If it contains relevant events, please answer "relevant" and list them as verbs only (no explanation).\\ 

\midrule

\cellcolor{gray!20} Prompt 2: Get implicit events & Your task is to find 5 implicit events in \{context\_paragraph\}. Explicit events have already been identified in \{explicit\_events\}. Implicit events are events that are not explicitly written in the paragraph, but that the reader understands happened. \newline Here is an example: \{example\}.\newline  Return the events as verbs, in a numbered list, followed by 2 explicit events that are the closest to the start and end time of the implicit event.\\ 

\midrule

\cellcolor{gray!20} Prompt 3: Construct sentences & Context Paragraph: \{context\_paragraph\}
Implicit and Explicit Events: \{instance\}
Your task is to construct pairs of events initially extracted from the \{context\_paragraph\}: events have already been extracted and are contained in \{instance\}. One instance includes one implicit event and two explicit events.
Construct 3 sentences that demonstrate the temporal relationship between the events. Use phrases like "Event A precedes Event B," "Event A is simultaneous with Event B", "Event A follows Event B" or "Event A happens after Event B" to indicate the temporal order. \newline Here is an example: \{example\}.\\ 
\bottomrule

\end{tabular}}
\caption{Prompts used for generating pre-labeled instances of \corpus.}
\label{table:datagen}
\end{table*}

\section{Prompts for Evaluation}\label{app: prompts_eval}
See Table \ref{table:prompts_eval} for the evaluation prompts used for various settings, and Tables \ref{table:lextime_eval} and \ref{table:tracie_eval} for examples from \corpus\xspace and \textsc{Tracie}, respectively.

\begin{table*}[t]
\centering
 
\resizebox{\textwidth}{!}{
\begin{tabular}{lp{20cm}}

\textbf{Description} & \textbf{Prompt} \\ 
\midrule

\cellcolor{gray!20} Zero-shot & 
Given the context, the task is to answer the query with "yes" or "no". \newline
context: "\{context\}" \newline
query: "\{query\}" \newline
answer: \\ 

\midrule

\cellcolor{gray!20} One-shot & 
Given the context, the task is to answer the query with "yes" or "no". \newline
example: \{example1\} \newline
context: "\{context\}" \newline
query: "\{query\}" \newline
answer: \\ 

\midrule

\cellcolor{gray!20} Few-shot & 
Given the context, the task is to answer the query with "yes" or "no". \newline
example 1: \{example1\} \newline
example 2: \{example2\} \newline
example 3: \{example3\} \newline
context: "\{context\}" \newline
query: "\{query\}" \newline
answer: \\ 

\midrule

\cellcolor{gray!20} CoT one-shot & 
Given the context, the task is to answer the query with "yes" or "no". \newline
Let's reason through it step-by-step: \newline
example: \{example1\_cot\} \newline
context: "\{context\}" \newline
query: "\{query\}" \newline
answer: \\

\midrule

\cellcolor{gray!20} CoT few-shot & 
Given the context, the task is to answer the query with "yes" or "no" .\newline 
Let's reason through it step-by-step: \newline
example 1: \{example1\_cot\} \newline
example 2: \{example2\_cot\} \newline
example 3: \{example3\_cot\} \newline
context: \{context\} \newline
query: \{query\} \newline
answer: \\ 

\bottomrule
\end{tabular}}
\caption{Prompt templates for evaluation, on all five types of prompts used: zero-shot, one-shot, few-shot, CoT one-shot, CoT few-shot. These templates are used for evaluation on both \corpus\xspace and \textsc{Tracie}, examples used differ and are extracted from each dataset respectively.}
\label{table:prompts_eval}
\end{table*}

\begin{table*}[t]
\centering
 
\resizebox{\textwidth}{!}{
\begin{tabular}{lp{20cm}}

\textbf{Description} & \textbf{Examples} \\ 
\midrule

\cellcolor{gray!20} Example 1 & 
context: "Before assigning the plaintiff to work, the defendant hired employees, including the plaintiff. The plaintiff was hired as a trackman."

query: "Does hiring employees happen before assigning the plaintiff to work?"

answer: "yes"

\\ 

\midrule

\cellcolor{gray!20} Example 2 & 
context: "ms. X advised ms. Y to issue the written discipline and to speak with mr. R (as well as the other roll tunnel operators) regarding the defaced guidelines. accordingly, ms. Y, along with John Doe (the union steward) approached mr. R to present him with the disciplinary warning and to discuss the defacement of the posted guidelines. mr. R was on his clamp truck and ms. Y advised that mr. R that she needed to speak with him. without stopping, mr. R told her that he did not have time to speak with her (his own supervisor) because he was working and drove by her."

query: "Does ms. X learning about the defaced guidelines precedes her advising ms. Y to issue the written discipline?"

answer: "yes" \\ 

\midrule

\cellcolor{gray!20} Example 3 & 
context: "rather than pursue the grievance through the contracts grievance and arbitration procedure, the union directed hospital employees to refuse to pick up extra shifts beyond their normal scheduled work. specifically, prior to the dispute over employees obligation to work holidays, employees would regularly and voluntarily pick up extra shifts beyond their normal scheduled work. but once the dispute over holiday work arose, employees stopped picking up shifts beyond their normal scheduled work. after the aforementioned dispute arose, a member of the union and a hospital employee provided to a member of the hospitals administration a text message stating, in part: per union unless you had originally picked up ot or st when it was first offered we should be giving back those hours for this coming week and they are asking us not to pick up anything through the end of may and the first 2 weeks of june because they still are not allowing people to take vacation on a holiday unless they have adequate staffing. upon information and belief, the references to ot and st in the text message mean overtime and straight time, respectively. the union members concerted refusal to pick up extra shifts is a violation of the contracts and has caused the hospital great hardship in achieving adequate staffing for the hospitals operations. as a result of the union and its members conduct, the hospital was forced to hire so-called traveler nurses and to pay an extraordinary premium, in the millions of dollars, for adequate staffing compared with the cost to the hospital had the union not directed its members to engage in this concerted effort and had the union members picked up extra shifts as they had in the past."

        query: "Does the hospital being forced to hire traveler nurses and pay an extraordinary premium precedes the union members' concerted refusal to pick up extra shifts?"

        answer: "no" \\ 

\midrule

\cellcolor{gray!20} Example 1 CoT & 
context: "Before assigning the plaintiff to work, the defendant hired employees, including the plaintiff. The plaintiff was hired as a trackman."

        query: "Does hiring employees happens before assigning the plaintiff to work?"

        answer:"
        Reasoning:
        1. The defendant hired employees, including the plaintiff.
        2. The plaintiff was hired as a trackman.
        3. The employees including the plaintiff were all assigned to work.
        
        The correct answer is: "yes"
 \\

\midrule

\cellcolor{gray!20} Example 2 CoT & 
context: "ms. X advised ms. Y to issue the written discipline and to speak with mr. R (as well as the other roll tunnel operators) regarding the defaced guidelines. accordingly, ms. Y, along with John Doe (the union steward) approached mr. R  to present him with the disciplinary warning and to discuss the defacement of the posted guidelines. mr. R was on his clamp truck and ms. Y advised that mr. R  that she needed to speak with him. without stopping, mr. R told her that he did not have time to speak with her (his own supervisor) because he was working and drove by her."

        query: "Does ms. X learning about the defaced guidelines precedes her advising ms. Y to issue the written discipline?"

        answer:"
        Reasoning:

        1. Ms. X learned about the defaced guidelines.
        2. She then advised Ms. Y to issue the written discipline.

        Since Ms. X's learning about the defaced guidelines happened before advising Ms. Y, the correct answer is "yes".
         \\ 

\midrule
\cellcolor{gray!20} Example 3 CoT & 
        context: "rather than pursue the grievance through the contracts grievance and arbitration procedure, the union directed hospital employees to refuse to pick up extra shifts beyond their normal scheduled work. specifically, prior to the dispute over employees obligation to work holidays, employees would regularly and voluntarily pick up extra shifts beyond their normal scheduled work. but once the dispute over holiday work arose, employees stopped picking up shifts beyond their normal scheduled work. after the aforementioned dispute arose, a member of the union and a hospital employee provided to a member of the hospitals administration a text message stating, in part: per union unless you had originally picked up ot or st when it was first offered we should be giving back those hours for this coming week and they are asking us not to pick up anything through the end of may and the first 2 weeks of june because they still are not allowing people to take vacation on a holiday unless they have adequate staffing[.] upon information and belief, the references to ot and st in the text message mean overtime and straight time, respectively. the union members concerted refusal to pick up extra shifts is a violation of the contracts and has caused the hospital great hardship in achieving adequate staffing for the hospitals operations. as a result of the union and its members conduct, the hospital was forced to hire so-called traveler nurses and to pay an extraordinary premium, in the millions of case 3:23-cv-00466 document 1 filed 04/13/23 page 5 of 10 6 dollars, for adequate staffing compared with the cost to the hospital had the union not directed its members to engage in this concerted effort and had the union members picked up extra shifts as they had in the past."

        query: "Does the hospital being forced to hire traveler nurses and pay an extraordinary premium precedes the union members' concerted refusal to pick up extra shifts?"

        answer:"
        Reasoning:
        
        1. The union members' concerted refusal to pick up extra shifts occurred as a result of the dispute.
        2. This refusal led to the hospital being forced to hire traveler nurses and pay an extraordinary premium.

        Since the hospital was forced to hire traveler nurses and pay a premium as a result of the union members' refusal, the correct answer is "no".
       
 \\

\bottomrule
\end{tabular}}
\caption{Examples used for one-shot and few-shot setting for evaluation on \corpus.}
\label{table:lextime_eval}
\end{table*}

\begin{table*}[t]
\centering
 
\resizebox{\textwidth}{!}{
\begin{tabular}{lp{20cm}}

\textbf{Description} & \textbf{Examples} \\ 
\midrule

\cellcolor{gray!20} Example 1 & 
        context: "My boss pulled me aside at work. I thought he was going to tell me Happy Birthday. Instead, he told me I wasn't needed today. I sulked home, not sure whether I was getting fired or not. When I arrived home, I walked into a surprise birthday party!"

        query: "Does the event i felt worried start after my boss told me i wasn't needed?"

        answer: "yes"

\\ 

\midrule
 
\cellcolor{gray!20} Example 2 & 
        context: "John got a new cat and was worried how it would take to the old one. He tried everything but the old cat would always act mean. One day the old cat was stuck in a box and the young cat helped it. Ever since they became best friends. "John was so happy with the result, he got another cat.""

        query: "Does the event john brought a cat home ends before the old cat stuck in a box?"

        answer: "yes" \\ 

\midrule

\cellcolor{gray!20} Example 3 & 
        context: "Javier has always wanted to try instant ramen. He decides that today is his day to try it. He goes to the store to buy a few packets. On his way home Javier is in a car accident. Javier never got to cook his instant ramen."

        query: "Does the event javier hasn't had instant ramen before. ends before he go to the store to buy a few packets ?"

        answer: "no" \\ 

\midrule

\cellcolor{gray!20} Example 1 CoT & 
        context: "My boss pulled me aside at work. I thought he was going to tell me Happy Birthday. Instead, he told me I wasn't needed today. I sulked home, not sure whether I was getting fired or not. When I arrived home, I walked into a surprise birthday party!"

        query: "Does the event i felt worried start after my boss told me i wasn't needed?"

        answer:"
        Reasoning:
        1. The boss informed the speaker that they weren't needed today.
        2. After hearing this news, the speaker felt worried about their job status.
        3. The feeling of worry is a response to the boss's statement.

        The correct answer is: "yes"
 \\

\midrule

\cellcolor{gray!20} Example 2 CoT & 
      context: "John got a new cat and was worried how it would take to the old one. He tried everything but the old cat would always act mean. One day the old cat was stuck in a box and the young cat helped it. Ever since they became best friends. "John was so happy with the result, he got another cat.""

        query: "Does the event john brought a cat home ends before the old cat stuck in a box?"

        answer:"
        Reasoning:

        1. John brought a new cat home and was concerned about how the old cat would react.
        2. The old cat initially acted mean towards the new cat.
        3. At some point, the old cat got stuck in a box.
        4. The event of bringing the new cat home occurs first and precedes the incident of the old cat getting stuck.

        Since  the correct answer is "yes". \\ 

\midrule
\cellcolor{gray!20} Example 3 CoT & 
        context: "Javier has always wanted to try instant ramen. He decides that today is his day to try it. He goes to the store to buy a few packets. On his way home Javier is in a car accident. Javier never got to cook his instant ramen."

        query: "Does the event javier hasn't had instant ramen before. ends before he go to the store to buy a few packets ?"

        answer:"
        Reasoning:
        
        1. The context states that Javier has always wanted to try instant ramen, indicating he has not had it before.
        2. Javier decides to try instant ramen today.
        3. He goes to the store to buy instant ramen, but he has not yet cooked or eaten it.
        4. Therefore, the event of not having instant ramen continues until he purchases the packets and cooks, meaning it does not end before he goes to the store.

        Since, the correct answer is "no".
 \\

\bottomrule
\end{tabular}}
\caption{Examples used for one-shot and few-shot setting for evaluation on \textsc{Tracie}.}
\label{table:tracie_eval}
\end{table*}

\end{document}